\documentclass[sigconf]{acmart}

\renewcommand\footnotetextcopyrightpermission[1]{} 
\usepackage{tikz} 
\usepackage{latexsym}
\usepackage{graphicx}
\usepackage{caption}
\usepackage{multirow}
\usepackage{tabularx}
\usepackage{url}
\usepackage{color}
\usetikzlibrary{shapes,arrows} 

\usepackage{times}
\usepackage{txfonts}
\newcommand{\argmax}{\mathop{\rm arg~max}\limits}

\usepackage{CJKutf8}

\begin{document}

\title{Summarizing Utterances from Japanese Assembly Minutes\\ using Political Sentence-BERT-based Method\\
for QA Lab-PoliInfo-2 Task of NTCIR-15}

\author{Daiki Shirafuji}
\affiliation{Graduate School of Information Science and Technology, Hokkaido University, Japan}
\email{d_shirafuji@ist.hokudai.ac.jp}
\author{Hiromichi Kameya}
\affiliation{School of Engineering, Hokkaido University, Japan}
\email{hkameya@eis.hokudai.ac.jp}
\author{Rafal Rzepka}
\affiliation{Faculty of Information Science and Technology,\\ Hokkaido University, Japan}
\email{rzepka@ist.hokudai.ac.jp}
\author{Kenji Araki}
\affiliation{Faculty of Information Science and Technology,\\ Hokkaido University, Japan}
\email{araki@ist.hokudai.ac.jp}

\begin{abstract}
There are many discussions held during political meetings, and a large number of utterances for various topics is included in their transcripts.
We need to read all of them if we want to follow speakers’ intentions or opinions about a given topic.
To avoid such a costly and time-consuming process to grasp often longish discussions, NLP researchers work on generating concise summaries of utterances.
Summarization subtask in QA Lab-PoliInfo-2 task of the NTCIR-15 addresses this problem for Japanese utterances in assembly minutes, and our team (SKRA) participated in this subtask.
As a first step for summarizing utterances, we created a new pre-trained sentence embedding model, i.e. the Japanese Political Sentence-BERT.
With this model, we summarize utterances without labelled data.
This paper describes our approach to solving the task and discusses its results.
\end{abstract}

\maketitle
\pagestyle{plain} 

\section*{Team Name}
SKRA

\section*{Subtasks}
Dialog Summarization task (Japanese)

\section{Introduction}

\begin{CJK}{UTF8}{min}
\begin{table}[t]
    \centering
    \begin{tabular}{|c|p{24em}|}
        \hline
        ID & \multicolumn{1}{c|}{Utterance} \\ \hline\hline
        1 &
        平成二十三年第二回都議会定例会の開会に当たりまして、都政運営に対する所信の一端を申し述べ、都議会の皆様と都民の皆様のご理解、ご協力を得たいと思います。
        (Before beginning the 2nd regular session of Tokyo Metropolitan Assembly of 2011, I would like to express my opinion on the management of the Tokyo metropolitan government and I want the members of the Tokyo Metropolitan Assembly and the people living in Tokyo for cooperation.)
        \\ \hline
        2 &
        戦後、我々は、アメリカ依存の平和に安住しながら繁栄を謳歌し、かつてない物質的な豊かさと引きかえに、日本人としての価値の基軸を失ってまいりました。
        (After the war, we enjoyed our prosperity with the peace which depends on the U.S. In exchange for unprecedented material wealth, we have lost the foundation of our values as Japanese people.)
        \\ \hline
    \end{tabular}
    \captionsetup{format=hang}
    \caption{Examples of utterance from the Tokyo Metropolitan Assembly Minutes dataset}
    \label{tab:example}
\end{table}
\end{CJK}

Researchers have recently been focusing on summarizing meetings \cite{chen2017semantic,verberne2020query,li2019keep}.
Li et al. \cite{li2019keep} work on summarizing meetings with multiple participants, which is significantly different from single documents such as news articles or reviews because a statement from a meeting participant may differ from statements of others.
These studies aim to summarize many utterances from different meeting participants into concise sentences considering contradictions between their utterances.
Tokyo Metropolitan Assembly Minutes dataset (Tokyo minutes dataset) is similar to the data in the aspect that several people share their opinions or ask questions to other participants (examples are shown in Table \ref{tab:example}).
The most different facet of this minute dataset from other discussions is the importance of who said what.
In the Tokyo minutes dataset, local politicians discuss on political topics, which are related to the management of the Tokyo metropolitan government.
They are usually required to be consistent in their statements and to follow their parties' stance to a given topic during a meeting.
The consistency of these statements will influence how Tokyo citizens will choose the candidates in the following Tokyo assembly elections.
Therefore, who said what is more important in Tokyo minutes dataset than in other types of usual meetings.

Kimura et al. \shortcite{kimura2019overview} were first to work on unique aspects of Tokyo minutes dataset and proposed three tasks, i.e. Segmentation, Summarization, and Classification.
They also organized QA Lab-PoliInfo task including those tasks during NTCIR-14 conference, and ask participants to solve them.
In the Segmentation task, participants extract the relevant sentences from a set of assembly utterances related to one summary.
In the Summarization task, its participants summarize utterances.
This is the subtask we address in this paper.
The purpose of the Classification task is to classify assembly utterances into three stances (pro, con, or neutral) for a given topic (e.g. ``The Tsukiji Market should move to Toyosu.'').
These three tasks help to quickly grasp the content of assembly meetings.
Summarization task will be more useful than other two tasks in order to promptly understand what politicians declare during a meeting because we can read a gist of their utterances at a glance.
Therefore, in this paper, we propose a method to summarize assembly transcripts.

\begin{CJK}{UTF8}{min}
\begin{table}[t]
    \centering
    \begin{tabular}{|p{4em}||p{20em}|}
        \hline
        Topic & 
        避難所の自立電源確保せよ
        (Tokyo metropolitan government should ensure self-sustaining power supply for evacuation shelters)
        \\ \hline
        Subtopic&
        避難所の自立電源
        (Self-sustaining power for evacuation shelters)
        \\ \hline
        Question Speaker &
        山口拓（立憲・民主） 
        (Taku Yamaguchi from Constitutional Democratic Party)
        \\ \hline
        Answer Speaker & 知事 (Tokyo Governor)\\ \hline
        Question Summary & 長期停電や平常時の温室効果ガス削減に役立つ電源確保の推進を。
        (We should promote securing power supplies to help reduce greenhouse gases during long-term power outages and during normal operation.)
        \\ \hline
        Answer Summary & 補正予算では、庁有車を災害時に非常用電源としても活用できる外部給電器の配備等を提案。引き続き非常時の電源確保を図る。 
        (In the supplementary budget, we proposed the deployment of an external power supply unit that can be used as an emergency power source in the case of a disaster for agency-owned vehicles. We will continue to work to secure power sources for emergency situations. )\\ \hline
    \end{tabular}
    \captionsetup{format=hang}
    \caption{Entry example from \textit{Togikai-dayori}, a portal providing summaries of Tokyo Metropolitan Assembly}
    \label{tab:togikaidayori}
\end{table}
\end{CJK}

Tokyo Metropolitan Assembly provides summaries for each meeting via a portal named ``\textit{Togikai-dayori}\footnote{You can access in \url{https://www.gikai.metro.tokyo.jp/newsletter/}.}'' (Tokyo Metropolitan Assembly Newsletters).
A summary originating from the site consists of a topic, a subtopic\footnote{A topic is the main theme of a meeting, and a subtopic is one of issues for the meeting discussion within a topic.}, an asker (Question Speaker), a replier (Answer Speaker), and short gists of questions and answers (see Table \ref{tab:togikaidayori}).
Although reading all utterances is possible, we can easily grasp what politicians claim from these summaries.
However, creating these summaries from utterances for each meeting is time-consuming and costly, thus there is a need to explore methods to automatically summarize them.
In addition to approaches for summarizing utterances, we have to consider adapting summarization techniques for Tokyo minutes dataset to be applicable to transcripts originating from other regions because only a few prefectures create their own summaries.
Therefore, we examine summarization methods without reference summaries.
We create a pre-trained model for sentence embeddings, and compute a similarity between each utterance and a given topic and a similarity between each utterance and a subtopic.
Then we calculate MMR-based function and extract utterances with the highest score.

In short, the main contributions of this article are:
\begin{enumerate}
\item Proposing Japanese Political Sentence-BERT;
\item Adapting an embedding-based unsupervised key-phrase extraction, EmbedRank++, to summarization.
\end{enumerate}

\section{Related Work}

\subsection{Summarization}
Almost all of standard non-neural summarization methods are based on extractive summarization
\cite{carbonell1998use,erkan2004lexrank,radev2004centroid,mihalcea2004textrank}.
Although researchers have proposed many neural methods to text summarization, those non-neural ones are still used for comparison \cite{lee2020experiments}.
In particular, Maximal Marginal Relevance (MMR) \cite{carbonell1998use}, LexRank \cite{erkan2004lexrank} and TextRank \cite{mihalcea2004textrank}  are typically the most frequently used methods.
MMR can be adapted to various methods, also in neural network-based methods because it is easy to be applied to score texts \cite{bennani2018simple,lebanoff2018adapting,fabbri2019multi}.
One of these algorithms, EmbedRank \cite{bennani2018simple}, is an unsupervised algorithm with sentence embeddings capable of extracting a key phrase from documents.
In addition, these MMR-based methods are easy to adapt not only to single document summarization (SDS), but also to multi-document summarization (MDS).

There are also abstractive summarization techniques for MDS with small datasets \cite{banerjee2016multi,lebanoff2018adapting}, which are recently based on deep neural networks \cite{gehrmann2018bottom,fabbri2019multi}.
Most recently, language models dedicated to summarization have also been developed, for example BERTSUM, PEGASUS and MARGE \cite{liu2019fine,zhang2019pegasus,lewis2020pre}.
However, Japanese language models for summarization do not exist to the best of the authors' knowledge.

\subsection{Sentence Embedding Method}
Word embedding-based methods recently became common tools in NLP field.
One of the most well-known algorithms, word2vec \cite{mikolov2013efficient}
is a pre-trained skip-gram or Continuous Bag-of-Words (CBOW) model to represent words as vectors.
Inspired by word2vec, pre-trained word embedding models have followed, for example Global Vectors (GloVe) \cite{pennington-etal-2014-glove} and fastText \cite{bojanowski2017enriching}.
In order to pre-train high quality representations, models need to capture semantics and different meanings the word can represent in different contexts.
To address these aspects, Embedding from Language Models (ELMo) \cite{Peters:2018} has been proposed.
This model uses word representations from a bi-directional LSTM (BiLSTM) trained with a language model (LM) on a massive data.

Following pre-trained word embedding models, sentence / document embedding ones have been developed.
For example, Skip-Thought Vector \cite{kiros2015skip} is based on a sentence encoder that predicts the surrounding sentences of a given sentence.
Skip-Thought is based on an encoder-decoder model, its encoder maps words to a sentence vector and its decoder generates the surrounding sentences.

Contrasting with Skip-Thought, InferSent \cite{conneau-EtAl:2017:EMNLP2017} is trained on labelled data, namely Stanford Natural Language Inference (SNLI) dataset \cite{bowman2015large}, for the sentence embeddings.

Sent2Vec \cite{pgj2017unsup} is an unsupervised objective to train distributed representations of sentences.
This model is inspired by CBOW, and it trains and infers sentence embeddings.

Doc2vec \cite{le2014distributed} is an algorithm based on Distributed Memory Model of Paragraph Vectors (PV-DM) or Distributed Bag of Words version of Paragraph Vector (PV-DBOW) to embed documents.

Universal Sentence Encoder (USE) \cite{cer2018universal} is a convolutional neural network-based method to embed sentences proposed by Cer et al. \cite{cer2018universal}.
For their method they implement Deep Averaging Network (DAN) and Transformer-based models.
In addition, they create multilingual USE models which cover 16 languages including Japanese \cite{yang2019multilingual,chidambaram2019learning}.

These above-mentioned sentence / document embedding models are available also in Japanese in addition to English.
However, most of recent sentence embedding models do not yet support Japanese language.
Sentence Transformers \cite{reimers-2019-sentence-bert} are sentence embedding models for English texts based on siamese / triplet networks \cite{hadsell2006dimensionality,hoffer2015deep}.
Reimers and Gurevych \cite{reimers-2019-sentence-bert} insert language models (BERT \cite{devlin2018bert}, RoBERTa \cite{liu2019roberta}, or XLM-RoBERTa \cite{conneau2019unsupervised}) into the network of siamese / triplet networks.
They call BERT-based siamese / triplet networks as ``Sentence-BERT.''
Although Sentence-BERT is available in English, to the best of our knowledge, no such model exists for Japanese political text.

In this research, we introduce Japanese Political Sentence-BERT, and test it with the political domain training on Japanese political utterances in Tokyo minutes dataset.

\section{Japanese Political Sentence-BERT}
In this section,
we describe how we create Japanese Political Sentence-BERT
with the given Tokyo minutes dataset provided by the task organizers \cite{poliinfo-2020-overview}.

\subsection{Data}

\begin{CJK}{UTF8}{min}
\begin{table}[t]
    \centering
    \begin{tabular}{|c|p{20.5em}|}
        \hline
        triplet & \multicolumn{1}{c|}{Utterance}\\ \hline\hline
        Target &
        このため、都は、用地取得の体制強化や都有地の活用など、一歩踏み込んだ支援メニューを示しまして、区の取り組みを促してまいりました。
        (Therefore, the Tokyo Metropolitan Government has presented solutions, such as strengthening the system for land acquisition and utilizing metropolitan-owned land, in order to encourage the city's efforts.)
        \\ \hline
        Positive &
        この結果、コア事業に位置づけた大田区糀谷駅前地区の再開発の工事が始まり、荒川区の都営住宅跡地において、生活道路整備のための従前居住者住宅の建設などが進んでおります。
        (As a result, work has begun on the redevelopment of the Kojiya-mae station area in Ota Ward, which was positioned as a core project, and the construction of former residents' houses on the site of a former municipal housing complex in Arakawa Ward is under construction in order to improve roads for daily life.)
        \\ \hline
        Negative & 
        次に、八ッ場ダムの治水効果でございますが、ただいまお答えした八つの洪水について、日本学術会議が妥当と判断した計算モデルを採用して、八ッ場ダムによる効果量を試算しております。
        (Next, regarding the flood control effects of the Yamba Dam, we estimated the effects of this dam for the eight flooding events mentioned above by using a simulation model determined to be appropriate by the Science Council of Japan.)
        \\ \hline
    \end{tabular}
    \captionsetup{format=hang}
    \caption{An example of [target sentence, positive sentence, negative sentence] triplet from Tokyo Metropolitan Assembly Minutes dataset}
    \label{tab:triplet}
\end{table}
\end{CJK}

In this section, we describe the data for pre-training Sentence-BERT to be used for Japanese political sentences.

We utilize utterances from Tokyo Metropolitan Assembly Minutes dataset (Tokyo minutes dataset) \cite{poliinfo-2020-overview}.
In the dataset, the utterances are listed chronologically (see Table \ref{tab:example} for the examples).
The IDs refer to time order, so after the utterance ID\_1 is spoken followed by the utterance ID\_2.
In addition to the utterances, there is some other information, for example the date, or the names of speakers.

To utilize this dataset for creating Japanese Sentence-BERT, we assumed that sentences that are adjacent to each other in terms of time series are semantically related.
We first compute an embedding of each utterance using Japanese NLP library, GiNZA\footnote{\url{https://github.com/megagonlabs/ginza}.}.
With these sentence embeddings, we calculate the cosine similarity between an utterance (which we call a ``target sentence'') and the utterance that follows it.
When the cosine similarity is 0.5 or more and the speaker is the same,
we regard a given pair as ``related utterances'' and define the extracted utterance after the target sentence as a ``positive sentence.''

In the next step, we retrieve a non-similar utterance following the above-mentioned target sentence from the data.
We extract a non-similar utterance with the following steps:
\begin{enumerate}
\item take an utterance which is spoken on a different day or in a different  meeting;
\item compute the sentence embedding of the extracted utterance with GiNZA;
\item calculate cosine similarity between the target sentence and the extracted utterance;
\item repeat steps 1-3 until the algorithm finds an utterance whose cosine similarity with the target sentence is 0.9 or less.
\end{enumerate}
After the fourth step, we finally extract one utterance which is dissimilar to the target sentence, and define it as a ``negative sentence.''
Note that thresholds (0.9 and 0.5) for sentence extraction steps are manually decided.

With these three types of sentences, we create a triplet, i.e. [target sentence, positive sentence, negative sentence] (see an example shown in Table \ref{tab:triplet}).
In the end, 27,078 triplets of [target sentence, positive sentence, negative sentence] were created.
These triplets are divided into 21,662 (80\%) for training, 2,708 (10\%) for development, and 2,708 (10\%) for testing.

\subsection{Model}

\begin{figure}[t]
\centering
\includegraphics[width=7cm]{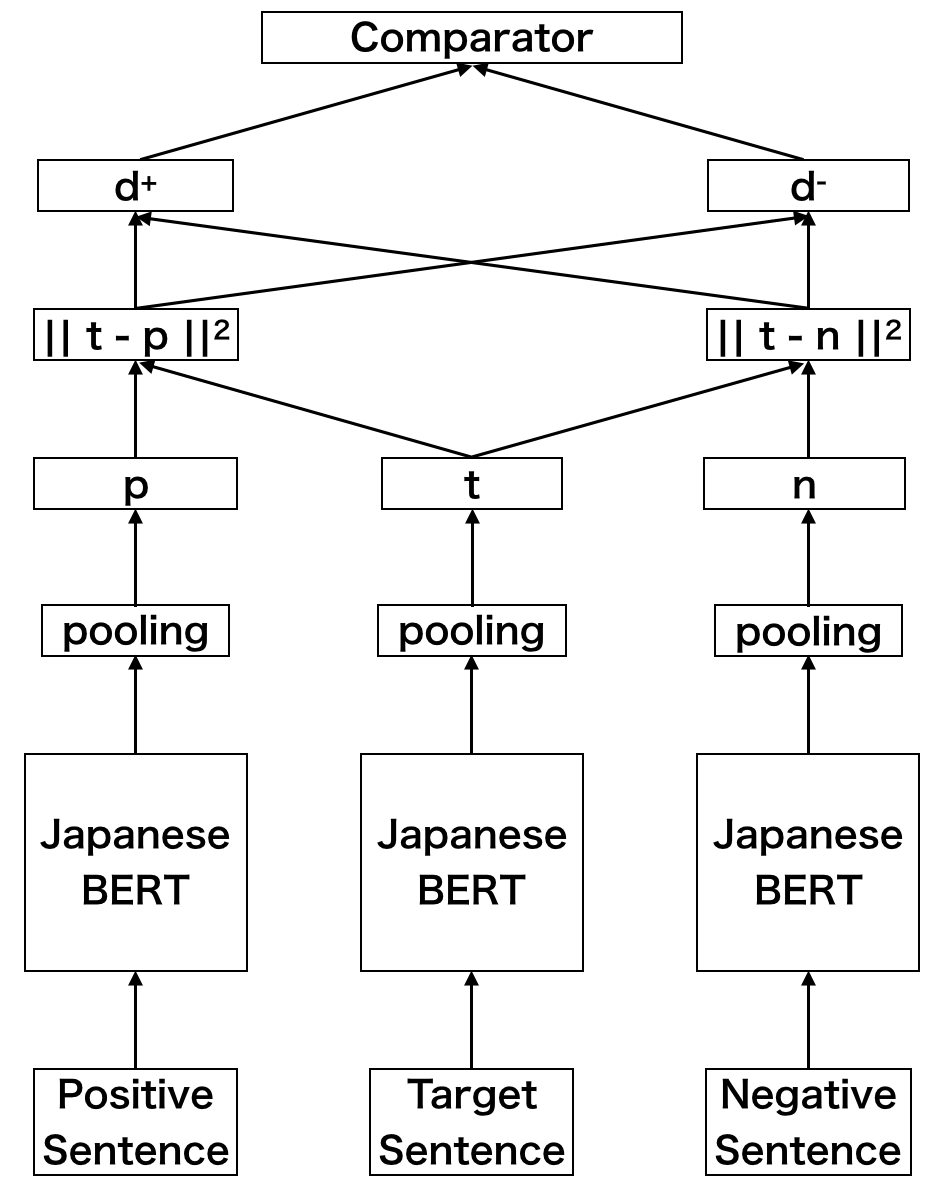}
\captionsetup{format=hang}
\caption{Creating Japanese Political Sentence-BERT with the triplet loss function}
\label{figure:Triplet}
\end{figure}


The pooling operation is added to Sentence-BERT model after the output from BERT in order to derive a fixed sized sentence embedding.
In our model, we adopt MEAN pooling, not MAX or CLS, because Reimers and Gurevych noticed that MEAN pooling is better than two other methods \cite{reimers-2019-sentence-bert}.
In order to fine-tune BERT on our dataset, we create a triplet network \cite{hoffer2015deep} to update the weights that make the computed utterance embeddings more semantically meaningful and the difference between a cosine similarity of a target sentence and a positive sentence and a cosine similarity of a target sentence and negative sentence becomes larger.

For a triplet network, the triplet loss function is defined.
Assuming we have three sentences; a target sentence $s_t$, a positive sentence $s_p$, and a negative sentence $s_n$, the triplet loss function tunes a triplet network to maximize the difference between $d_{\mbox{+}}$ and $d_-$ which are defined with Equation (\ref{distance}):

\begin{equation} \label{distance}
\begin{split}
d_{\mbox{+}} & \coloneqq \frac{||e^{f(s_t)-f(s_p)}||_2}{||e^{f(s_t)-f(s_p)}||_2 \mbox{+} ||e^{f(s_t)-f(s_n)}||_2},\\
d_{-} & \coloneqq \frac{||e^{f(s_t)-f(s_n)}||_2}{||e^{f(s_t)-f(s_p)}||_2 \mbox{+} ||e^{f(s_t)-f(s_n)}||_2},
\end{split}
\end{equation}

where $f$ is an embedding function.
The triplet loss is defined by the following Equation (\ref{TripletLoss}):

\begin{equation} \label{TripletLoss}
\begin{split}
L\mbox{(}d_{\mbox{+}}, d_-\mbox{)} & \, \mbox{=} \, ||d_{\mbox{+}}, d_{-}-1||_{2}^{2}\\
& \,\mbox{=}\, const*d_+^2.
\end{split}
\end{equation}

We normalize to make $d_{\mbox{+}} \mbox{+} d_- \mbox{=} 1$,
so $d_-$ should be $1$ when $d_{\mbox{+}} \mbox{=} 0$.
Therefore, the triplet loss can be also described by Equation (\ref{TripletLoss2}):
\begin{equation} \label{TripletLoss2}
\begin{split}
L\mbox{(}s_t, s_p, s_n\mbox{)} & \, \mbox{=} \, d\mbox{(}f\mbox{(}s_t\mbox{)},f\mbox{(}s_p\mbox{)}\mbox{)}
\mbox{+} \mbox{(}1-d\mbox{(}f\mbox{(}s_t\mbox{)}-f\mbox{(}s_n\mbox{)}\mbox{)}\mbox{)}\\
& \, \mbox{=} \, 2d\mbox{(}f\mbox{(}s_t\mbox{)}-f\mbox{(}s_p\mbox{)}\mbox{)},
\end{split}
\end{equation}
where $d$ is Euclidean distance function. This Equation (\ref{TripletLoss2}) shows that $d_{\mbox{+}}$ becomes minimized when $d_-$ is maximized.

In our model, we insert the pre-trained Japanese BERT\footnote{We use the BERT-base model (12-layer, 768-hidden, 12-heads, and 110M parameters) with Whole Word Masking released by Tohoku University. The model is available at \url{https://github.com/huggingface/transformers}.} into the branches of the triplet network,
and perform pre-training the triplet network on the triplet data described in Section 3.1. 
Again, we regard a target sentence as $s_t$, a positive sentence as $s_p$, and a negative sentence as $s_n$.
We set the margin for triplet loss $\epsilon$ to $1$ following the Sentence-BERT model for English \cite{reimers-2019-sentence-bert}.

The overview of creating Japanese Sentence-BERT is shown in Figure \ref{figure:Triplet}.
The ``Comparator'' in Figure \ref{figure:Triplet}  refers to the comparing function to identify which sentence, $s_p$ or $s_n$, is positive / negative.

\subsection{Training Details}
We train Sentence-BERT on the triplet data as described in Section 3.1.
As mentioned before, our data consists of 27,078 triplets ([target sentence, positive sentence, negative sentence]).
These triplets are divided into 21,662 (80\%) for the train data, 2,708 (10\%) for the development data, and 2,708 (10\%) for the test data.
We use a batch size of 16, Adam optimizer with learning rate 2e-5, and a linear learning rate warm up over 10\% of the train data as in the Sentence-BERT model for English \cite{reimers-2019-sentence-bert}.
We train a Japanese triplet BERT network for three epochs with the above-mentioned hyper-parameters.
The pooling strategy is MEAN.

Our dataset we train Sentence-BERT on comes from Tokyo Metropolitan Assembly Minutes dataset consisting of political utterances.
Therefore, the acquired Sentence-BERT model specializes in political related sentences, hence we name the model ``Japanese Political Sentence-BERT.''

Note that our training details in the most part follows Sentence Transformers algorithms\footnote{Sentence Transformers are available in \url{https://github.com/UKPLab/sentence-transformers}.
The version of the library we utilize is 0.3.2.} except for the epoch number.

\subsection{Comparing Method and Metrics}
We utilize Universal Sentence Encoder (USE) \cite{cer2018universal} for the comparison with our Japanese Political Sentence-BERT for a sentence embedding function $f$.
USE is one of the most common methods to embed Japanese sentences into vectors as mentioned in Section 2.2.
To show the usability of our Sentence-BERT model, we perform tests with both algorithms.
The evaluation steps are as follows:
\begin{enumerate}
\item embed the sentences in a triplet of [target sentence, positive sentence, negative sentence] in test data with the BERT modules inside the triplet network;
\item compute cosine similarities between [target sentence, positive sentence] and [target sentence, negative sentence] from sentence embeddings;
\item calculate the difference between two acquired cosine similarities of [target sentence, positive sentence] and [target sentence, negative sentence], and evaluate the accuracy by counting how many of these differences are equal to zero or larger.
\end{enumerate}

We define the difference between two acquired cosine similarities of [target sentence, positive sentence] and [target sentence, negative sentence] as $\mathit{diff}$,
and $\mathit{diff}$ can be represented as Equation (\ref{Difference}):
\begin{equation} \label{Difference}
\mathit{diff} \,\mbox{=}\,
\mathit{CosSim}\mbox{(}f\mbox{(}s_t\mbox{)}, f\mbox{(}s_p\mbox{)}\mbox{)}
- \mathit{CosSim}\mbox{(}f\mbox{(}s_t\mbox{)}, f\mbox{(}s_n\mbox{)}\mbox{)}.
\end{equation}
We evaluate models also with a $\mathit{diff}$ function in addition to accuracy.
The larger $\mathit{diff}$ indicates that the model can identify a positive sentence and a negative sentence better.

The USE model which we use for a sentence embedding function $f$ is the Universal Sentence Encoder Multilingual Large module\footnote{The model is available in \url{https://tfhub.dev/google/universal-sentence-encoder-multilingual-large/1}.} \cite{yang2019multilingual,chidambaram2019learning}.
USE model uses a Convolutional Neural Network-based approach to embed sentences, and covers 16 languages including Japanese.
The length of its inputted text to USE is variable, and the dimension of an outputted sentence vector is 512.

\subsection{Evaluation Results}
\begin{table}[t]
    \centering
    \begin{tabular}{|c||r|r|}
        \hline
         & \multicolumn{1}{c|}{$\mathit{diff}$} & \multicolumn{1}{c|}{Accuracy} \\ \hline \hline
        USE & 0.2441 & 0.8674 \\ \hline
        JPSB (Proposed Model) & \textbf{0.3705} & \textbf{0.9849} \\ \hline
    \end{tabular}
    \captionsetup{format=hang}
    \caption{Results of USE and our proposed Japanese Political Sentence-BERT (JPSB) evaluated on the test data}
    \label{tab:result4SBERT}
\end{table}
This section compares the methods evaluated on the test data, and shows the results in Table \ref{tab:result4SBERT}.
Japanese Political Sentence-BERT outperforms the existing sentence embedding method (USE) by approximately 0.12 in accuracy.
Regarding $\mathit{diff}$, Japanese Political Sentence-BERT exceeds USE by around 0.13.
Both of these results show that our Sentence-BERT model represents a sentence embedding more effectively than the existing sentence embedding models.

\subsection{Considerations about Japanese Political Sentence-BERT}
To the best of authors' knowledge, this is the first attempt to implement Sentence-BERT for Japanese language.
Our model specializes in political sentences, which are the core of the QA Lab-PoliInfo-2 tasks at NTCIR-15 evaluation conference.
Japanese Political Sentence-BERT outperforms the USE approach by around 0.12 in accuracy and by around 0.13 in $\mathit{diff}$.
From this difference, it is sufficient to say our model represents sentence embeddings better than other existing models.

As the future work, we plan to create Japanese Sentence-BERT for general texts trained on Wikipedia data, and to examine the usefulness of Japanese Sentence-BERT and Japanese Political Sentence-BERT for other QA Lab-PoliInfo-2 subtasks, i.e. the stance classification task and the entity linking task.

\section{Summarization Subtask}
In this section, we describe the summarization subtask in QA Lab-PoliInfo-2 and results of our attempt to summarize utterances in  Tokyo Metropolitan Assembly Minutes dataset with Japanese Political Sentence-BERT described in Section 3.

\subsection{Data}
\begin{CJK}{UTF8}{min}
\begin{table}[t]
    \centering
    \begin{tabular}{|p{8.7em}||p{14em}|}
        \hline
        Answer ending line & 0 \\ \hline
        Answer length & 150 \\ \hline
        Answer speaker & 知事 (Tokyo Governer) \\ \hline
        Answer starting line & 0 \\ \hline
        Answer summary & None \\ \hline
        Date & 2013-02-26 \\ \hline
        ID & PoliInfo2-DialogSummarization-JA-Formal-Test-00001 \\ \hline
        Main topic & 新知事の東京の将来像を示せ<br>エネルギー需要側の政策進化を (Show us the future goals for Tokyo which you, the new governor, have. <br> You should improve the policy from the perspective of the energy demanding side.) \\ \hline
        Subtopic & 都政運営 (Administration of Tokyo Metropolitan Government) \\ \hline
        Meeting & 平成25年第1回定例会 (the 1st regular session in the year 2013) \\ \hline
        Prefecture & 東京都 (Tokyo) \\ \hline
        Question ending line & 0 \\ \hline
        Question length & 100 \\ \hline
        Question speaker & 酒井大史（民主党） (Daishi Sakai from Democratic Party) \\ \hline
        Question starting line & 0 \\ \hline
        Question summary & None \\ \hline
    \end{tabular}
    \captionsetup{format=hang}
    \caption{Example of the inputted test data for the summarization task}
    \label{tab:SumTestExample}
\end{table}
\end{CJK}

We experiment with the test data provided by the QA Lab-PoliInfo-2 task organizers \cite{poliinfo-2020-overview} for the summarization task.
In our method, we do not use training data provided by the task organizers.
As shown in the test data example in Table \ref{tab:SumTestExample}, there is some additional information regarding one question.
Each question must have at least one answer.
The important data components for our approach to the summarization task (see Section 4.3) are:
\begin{enumerate}
\item \textbf{Answer length} showing the number of character limits of the answer summary,
\item \textbf{Answer Speaker} referring the name of participant who answer to a given question,
\item \textbf{Main topic} being a theme for the meeting,
\item \textbf{Subtopic} being a category for the utterance spoken during the meeting,
\item \textbf{Question length} showing the number of character limits of the question summary,
\item \textbf{Question Speaker} referring the name of participant who asked a question.
\end{enumerate}
Both numbers of character limits are decided by the task organizers.
In the test data, 254 questions are included.

\subsection{Evaluation Metrics}
For evaluation, we used two summary evaluation metrics: recall score in ROUGE-1 \cite{lin2004rouge} for both methods (described in Section 4.3.1), and human evaluation for the Japanese Political Sentence-BERT-based model.
ROUGE is a recall-based metric for fixed-length summaries which is based on n-gram co-occurrence.
Regarding human evaluation, the task organizers evaluate our results with the following metrics:
\begin{itemize}
\item \textbf{Content}: evaluates output semantically. The range of scores are [X, 0, 1, 2], and the larger score, the better summary. Note that the label X means the output summary is a concise statement which summarizes utterances, but differs from the reference summary. The organizers evaluate this metric with two scores: X=2 (when the regarding the different content summaries as correct) and X=0 (regarding the different content summaries as incorrect).
\item \textbf{Well-formed}: evaluates the naturalness of an output. The range of scores is [0, 1, 2], and the larger score, the better summary;
\item \textbf{Non-twisted}: evaluates whether the outputted texts express the speaker's intention without distorting it. The range of scores is [0, 1, 2], and the larger score, the better summary;
\item \textbf{Sentence goodness}: evaluates whether an outputted text expresses a concise summary or not. The range of scores is [0, 1, 2], and the larger score, the better summary;
\item \textbf{Dialog goodness}: evaluates whether outputted texts express a concise summary for a set of questions and answers. The range of scores is [0, 1, 2], and again the larger score, the better summary.
\end{itemize}

The human evaluation score for all summaries is calculated with the sum of scores and their averages.
Each summary is scored by two of the summarization task participants chosen by the QA Lab-PoliInfo-2 organizers.

\subsection{Methods}
This section explains our approach for summarizing utterances with sentence embeddings.
We describe the summarization method in Section 4.3.1, and show two sentence embedding models which are used in the summarization method described in Section 4.3.2.

\subsubsection{Our Proposed Method}
We propose the sentence embedding model-based approach for the utterance summarization which is described below.
The sentence embedding models which we use for the method are described in Section 4.3.2.

We introduce a MMR-based extractive method, i.e. EmbedRank \cite{bennani2018simple}, to our algorithm.
EmbedRank is an unsupervised method with sentence embeddings for key word extraction.
Its creators also proposed EmbedRank++ in order to avoid extracting key phrases of the same meaning and acquire different words.
Both algorithms differ from traditional methods of key phrase extraction based on graph representations \cite{erkan2004lexrank,mihalcea2004textrank}, because they utilize sentence embeddings for extracting phrases similar to documents.
We adapt EmbedRank++ to extract sentences.
Moreover, to obtain more accurate sentence embeddings, which are described in Section 4.3.2, we use our Japanese Political Sentence-BERT model and Universal Sentence Encoder (USE) model, which are described in Section 4.3.2, instead of Doc2vec \cite{le2014distributed} and Sent2Vec \cite{pgj2017unsup} which were used for English texts \cite{bennani2018simple}.

In EmbedRank++, there are two parameters; \textit{key size} and $\lambda$.
\textit{Key size} defines the number of extracted key phrases.
When \textit{key size} equals 1, only one key phrase is extracted by EmbedRank++.
$\lambda$ is a parameter in the equation for Maximal Marginal Relevance (MMR) \cite{carbonell1998use} which is introduced in EmbedRank++, and controls the diversity of extracted words and the relevance between documents and extracted words.
With a given input query $Q$, the set $S$ represents documents that are selected as correct answers for $Q$.
$S$ is populated by computing MMR as described in Equation (\ref{MMR}),

\begin{equation} \label{MMR}
\begin{split}
MMR \coloneqq 
\argmax_{D_i \in R\backslash S}
\{ & \lambda*Sim_1 \left( D_i, Q \right) \\
   & -\left(1-\lambda\right) \max_{D_j \in S} Sim_2\left(D_i,D_j\right)\},
\end{split}
\end{equation}

where
$R$ is the ranked list of documents retrieved by an algorithm,
$S$ represents the subset of documents in $R$ which are already selected,
$D_i$ and $D_j$ are retrieved documents,
$Q$ is the averaged vector of inputted all documents
and
$Sim_1$ and $Sim_2$ are similarity functions.
Following the existing research on MMR \cite{carbonell1998use}, we set $\lambda$ to 0.5.
Using cosine similarity for the similarity functions, we compute MMR score between $D_i$ and $Q$ and between $D_i$ and $D_j$.

For the summarization task, we introduce two functions in addition to two terms in Equation (\ref{MMR}).
One is the cosine similarity between $D_i$ and a given main topic $MT$, and the other is also the cosine similarity between $D_i$ and a given subtopic $ST$.
MMR score for our method is calculated with the following Equation (\ref{MMRforSum}):

\begin{equation} \label{MMRforSum}
\begin{split}
MMR \, \mbox{=} \,
\argmax_{D_i \in R\backslash S}
\{ & k*\{0.5*\mathit{CosSim} \left( D_i,Q \right) \\
   & - 0.5*\max_{D_j \in S} \mathit{CosSim} \left(D_i,D_j\right)\} \\
   & \mbox{+} m * \mathit{CosSim} \left(D_i, MT\right) \\
   & \mbox{+} s * \mathit{CosSim} \left(D_i, ST\right)
   \},
\end{split}
\end{equation}

where $k$, $m$ and $s$ are parameters to assign weight to each similarity.
In order to obtain MMR score, Bennani-Smires et al. \shortcite{bennani2018simple} computed embeddings of documents with Doc2vec and Sent2Vec,
and we adapt Japanese Political Sentence-BERT and USE in this summarization task.

The parameter \textit{key size} is set up accordingly to the following Equation (\ref{KeySize}):
\begin{equation} \label{KeySize}
\begin{split}
key\,size\left(Answer\right) &\coloneqq \frac{Answer \, length}{50},\\
key\,size\left(Question\right) &\coloneqq \frac{Question \, length}{50}.
\end{split}
\end{equation}
The reason why we divide the lengths by 50 is that we assume the 50 characters may be the maximum length to express one content.
Other parameters $k$, $m$ and $s$ are set to $0.2$, $0.3$, and $0.5$, respectively.
These three parameter settings are decided to bring the total to 1.

\subsubsection{Sentence Embeddings for the proposed method}
We utilize the Japanese Political Sentence-BERT described in the Section 3 to the sentence embedding model in our above-mentioned approach (see Section 4.3.1).
We also adopt Universal Sentence Encoder (USE) as the comparison method, which is one of the common sentence embedding models and is available for Japanese language.
We examine sentence embeddings with both models, and summarize utterances using the method described in Section 4.3.1.

\subsection{Evaluation Results}
\begin{table}[t]
    \centering
    \begin{tabular}{|p{15em}||r|}
        \hline
         & \multicolumn{1}{c|}{ROUGE-1} \\ \hline \hline
        \multicolumn{1}{|c||}{USE-based EmbedRank++} & 0.0846 \\ \hline
        Japanese Political Sentence-BERT-based EmbedRank++ & 0.0696 \\ \hline
    \end{tabular}
    \captionsetup{format=hang}
    \caption{Results of USE-based and Japanese Political Sentence-BERT-based EmbedRank++ tested on the test data}
    \label{tab:result4ROUGE1}
\end{table}

\begin{table}[t]
    \centering
    \begin{tabular}{|c|c|c||r|}
        \hline
        \multicolumn{3}{|c||}{Human Evaluation} & \multicolumn{1}{c|}{Our Model} \\ \hline \hline
        \multicolumn{3}{|c||}{num of summaries} & 533 \\ \hline
        \multirow{4}{*}{Content} & \multirow{2}{*}{(X=2)} & sum & 184.5 \\ \cline{3-4}
        & & average & 0.346 \\ \cline{2-4}
        & \multirow{2}{*}{(X=0)} & sum & 135.5 \\ \cline{3-4}
        & & average & 0.254 \\ \hline
        \multicolumn{2}{|c|}{\multirow{2}{*}{Well-formed}} & sum & 888.5 \\ \cline{3-4}
        \multicolumn{2}{|c|}{} & average & 1.667 \\ \hline
        \multirow{4}{*}{Non-twisted} & \multirow{2}{*}{all} & sum & 184.5 \\ \cline{3-4}
        & & average & 0.346 \\ \cline{2-4}
        & \multirow{2}{*}{evaluable} & sum & 135.5 \\ \cline{3-4}
        & & average & 0.254 \\ \hline
        \multicolumn{2}{|c|}{\multirow{2}{*}{Sentence goodness}} & sum & 151.5 \\ \cline{3-4}
        \multicolumn{2}{|c|}{} & average & 0.284 \\ \hline
        \multicolumn{2}{|c|}{\multirow{2}{*}{Dialog goodness}} & sum & 43.0 \\ \cline{3-4}
        \multicolumn{2}{|c|}{} & average & 0.169 \\ \hline
    \end{tabular}
    \captionsetup{format=hang}
    \caption{Human evaluation results of Japanese Political Sentence-BERT-based EmbedRank++ tested on the test data}
    \label{tab:result4Human}
\end{table}

Table \ref{tab:result4ROUGE1} shows recall of ROUGE-1 scores,
and Table \ref{tab:result4Human} shows human evaluation results.
Note that the manual evaluation is conducted by the task organizers, so there are no results of USE-based approach.

Regarding ROUGE-1 scores, Sentence-BERT-based approach does not outperform USE-based one.
Except well-formed scores in the human evaluation metric, our method does not perform well when compared to other task participants (all results are shown in the task overview paper \cite{poliinfo-2020-overview}).

\subsection{Discussion for the Summarization Subtask}

\begin{CJK}{UTF8}{min}
\begin{table}[t]
    \centering
    \begin{tabular}{|p{4em}||p{20em}|}
        \hline
        Reference Summary \#1 &
        都の産業政策やインフラ整備との一体的取組、川崎港・横浜港との連携に加え、都が責任を持って経営に関わる体制を確保。
        \small{(We should ensure that the Tokyo Metropolitan Government takes responsibility for management of the Port of Kawasaki and the Port of Yokohama, in addition to coordinated efforts with Tokyo's industrial policies and infrastructure development and cooperation with the Port of Kawasaki and the Port of Yokohama.)}
        \\ \hline
        Output \#1 & 
        東京の産業政策やインフラ整備と一体的に取り組むことが求められております。/現場の実態を熟知した東京都が責任を持って港湾経営にかかわっていく体制を確保してまいります。
        \small{(We need to work at the same time on Tokyo's industrial policies and infrastructure development. We will ensure that the Tokyo Metropolitan Government, which is well versed in the actual situation on the ground, will take responsibility for port management.)}
        \\ \hline
        \hline
        Answer Summary \#2 & 都道について新たな整備目標示す。区市町村支援、国への財源拡充要求、電線事業者等との連携を強化し積極的に推進する。
        \small{(We will explain the new purposes for Tokyo metropolitan roads; actively promoting support for municipalities; demanding expansion of financial resources from the government; and strengthening cooperation with wire-line operators.)}
        \\ \hline
        Output \#2& 
        一層の支援に努めてまいります。/風格ある都市景観の形成と高度防災都市の実現を目指し世界に誇れる都市空間を創出してまいります。
        \small{(We will make every effort to support the city. We will create a world-class urban space with the aim of creating a distinctive urban landscape and a highly disaster-resistant city.)}
        \\ \hline
    \end{tabular}
    \captionsetup{format=hang}
    \caption{Example of the outputted summaries for the summarization subtask}
    \label{tab:SumTestExample}
\end{table}
\end{CJK}

From these results, we can find that our method does not improve the scores.
Furthermore, our Sentence-BERT model does not contribute to extracting accurate utterances for summarizing assembly minutes.
One of the main reasons why the method does not achieve high results is that we do not tune parameters in our model: $k$, $m$ and $s$.
We need to tune these three parameters with the train data provided by the task organizers, and acquire the set of parameters to maximize recall in ROUGE-1 score in the near future.
Considering the outputted results, shown in the Table 8, the outputted summary \#1 expresses the similar theme and contents to the reference summary.
On the other hand, the outputted summary \#2 cannot refer totally different theme from the reference summary.
This means that Equation (\ref{MMRforSum}) cannot control the related sentence to a given theme well.
We need to change parameters in Equation (\ref{MMRforSum}) so that the parameters for a similarity between a sentence and a given main topic and a similarity between a sentence and a given subtopic are increased.

Regarding the comparison of USE-based approach and Sentence-BERT-based one, it cannot be said that our pre-trained Japanese Political Sentence-BERT performs well in a downstream task, i.e. a summarization task.
Our model requires changing the hyper-parameters and the data.
We plan to implement Sentence-BERT modules trained on news corpus for political domain tasks.
In addition, for general NLP tasks, we will also train our model on Japanese Wikipedia.

\begin{CJK}{UTF8}{min}
There are some cases where the outputted texts were meaningless as a summary.
For example, ``（拍手）'' (annotation for hand clapping) or ``--'' were found in the output. We need to delete such noise in the pre-processing.
\end{CJK}

\section{Conclusions}
Our work was the first attempt to create Sentence-BERT for Japanese texts, and to adopt it to utterances included in the Japanese minute data summarization task.
This model outperformed the existing sentence embedding model USE by around 0.12 in accuracy.
Regarding the utterance summarization, we adopted MMR-based unsupervised approach, EmbedRank++, for utterance summarization, and our Japanese Sentence-BERT for sentence embeddings.
As comparison with the Sentence-BERT, we also utilized USE as a sentence embedding model for EmbedRank++.
Additionally, we changed the MMR function for EmbedRank++ and added two cosine similarity functions: one was a similarity between a given main topic and an utterance, and the other was a similarity between a given subtopic and an utterance.
Our method succeeded to output natural sentences for summaries.
However, in spite of algorithm improvements, we discovered that our EmbedRank++ method based on Japanese Political Sentence-BERT without parameter tuning does not contribute to generating accurate texts for summaries of utterances.
In the future, we plan to change parameters during the training, and use the parameters to acquire better results.
Regarding Sentence-BERT, our model did not perform well in this subtask, so we plan to pre-train it with other data, for example with news corpus or Wikipedia.

\bibliographystyle{ACM-Reference-Format}

\bibliography{ntcir15.bbl}

\end{document}